%% file: main.tex
\documentclass{INTERSPEECH2024}

\interspeechcameraready

\title{Speech Prefix-Tuning with RNNT Loss for Improving LLM Predictions}
\name{Murali Karthick Baskar, Andrew Rosenberg, Bhuvana Ramabhadran, Neeraj Gaur, Zhong Meng}
\address{
  Google Inc., USA}
\email{mkbaskar@google.com,rosenberg@google.com}

\input{inputfiles/packages}
\begin{document}

\maketitle
 
\begin{abstract}
\input{inputfiles/abstract}
\end{abstract}

\section{Introduction}
\input{inputfiles/intro}

\section{Methodology}
\input{inputfiles/methods}

\section{Experiments}
\input{inputfiles/experiments}

\section{Results and Discussion}
\input{inputfiles/discussion}

\section{Related works}
\input{inputfiles/related}

\section{Conclusions}
\input{inputfiles/conclusion}

\bibliographystyle{IEEEtran}
\bibliography{mybib}

\end{document}

%% file: inputfiles/packages.tex
\usepackage{booktabs}
\usepackage{amsmath}
\usepackage{pgfplots}
\usepackage{highlight}
\usepackage{xcolor,colortbl}
\usepackage{subfig}
\usepackage{multirow}
\pgfplotsset{compat=1.16}

%% file: inputfiles/abstract.tex
In this paper, we focus on addressing the constraints faced when applying LLMs to ASR. Recent works utilize prefixLM-type models, which directly apply speech as a prefix to LLMs for ASR. We have found that optimizing speech prefixes leads to better ASR performance and propose applying RNNT loss to perform speech prefix-tuning. This is a simple approach and does not increase the model complexity or alter the inference pipeline. We also propose language-based soft prompting to further improve with frozen LLMs. Empirical analysis on realtime testset from 10 Indic languages demonstrate that our proposed speech prefix-tuning yields improvements with both frozen and fine-tuned LLMs. Our recognition results on an average of 10 Indics show that the proposed prefix-tuning with RNNT loss results in a 12\% relative improvement in WER over the baseline with a fine-tuned LLM. Our proposed approches with the frozen LLM leads to a 31\% relative improvement over basic soft-prompting prefixLM.

%% file: inputfiles/intro.tex
Large language models (LLMs) are revolutionizing automatic speech recognition (ASR) research by addressing various types of prediction errors. PrefixLM is an LLM variant where the input text is accompanied by a prefix. This prefix can take the form of text~\cite{raffel2020exploring} or speech~\cite{fathullah2023prompting} 
or image~\cite{mokady2021clipcap} providing additional context for the model. When using speech tokens as prefixes (as in this work), PrefixLM learns to predict text autoregressively, mimicking an end-to-end ASR model~\cite{multe2e,kim2023prefix}. Previous work~\cite{embarassprefix} 
demonstrates that LLM performance improves with better speech encodings or prefix tokens extracted from self-supervised and supervised models. Scaling the speech encoder also enhances the use of speech prefixes~\cite{fathullah2023prompting,lakomkin2023end}, further improving the recognition ability of LLM models such as LLaMA~\cite{touvron2023llama}. PrefixLM with speech prefixes have been trained for multiple tasks, including speech recognition and speech translation~\cite{wang2023slm}. Notably, these approaches directly use pretrained LLMs without additional fine-tuning on the target task.

Prefix-tuning~\cite{li2021prefix} offers a lightweight alternative to fine-tuning, as it prepends a trainable token sequence to the text input. Optimizing only the prefix-related parameters adapts the model effectively to downstream tasks.  This technique is being incorporated into image and video-based prefixLM models as well. 
Cross-modal prefix-tuning~\cite{ma2022cpt} has been proposed for adapting multilingual translation models to bilingual speech translation tasks. 
While the final training objective is still only the LLM loss, a pre-trained speech encoder is used for adaptation across modalities.
While LLMs gain significant improvement in recognition performance, they still suffer from drawbacks such as higher insertions~\cite{ma2023can} and code-switching errors~\cite{hu2023massively}. The authors in~\cite{ma2023can} demonstrate that using LLMs for error correction for ASR helps to improve substitutions compared to RNNTs but increases insertion errors. A simple shallow fusion with a bi-directional LLM leads to code-switching between Mandarin and English when compared to an RNNT model~\cite{hu2023massively}. Despite these studies, our multilingual experiments also indicate that RNNTs do not suffer from insertions and code-switching to the same degree as LLM predictions. We attribute this behavior of RNNT to training with robust aligments and hypothesize that integrating it with LLM will lead to reduced hallucinations and better prediction. 
\begin{figure}[!ht]
    \centering
    \subfloat[a]{\includegraphics[width = 1.0in]{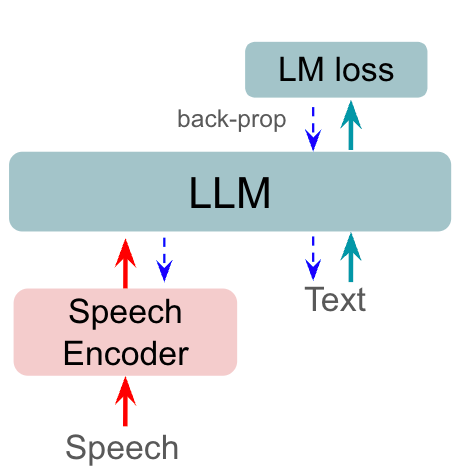}} 
    \subfloat[b]{\includegraphics[width = 1.2in]{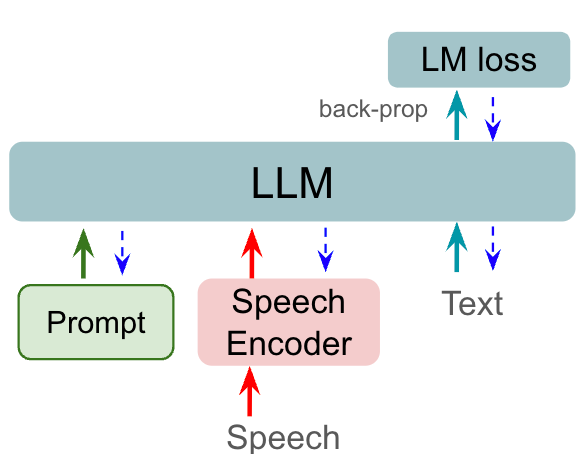}}\\
    \subfloat[c]{\includegraphics[width = 1.0in]{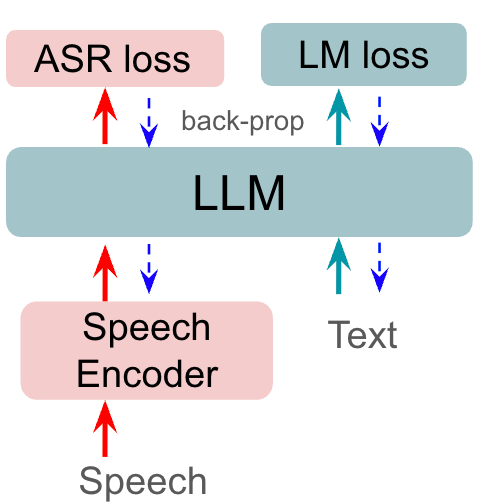}}
    \subfloat[d]{\includegraphics[width = 1.2in]{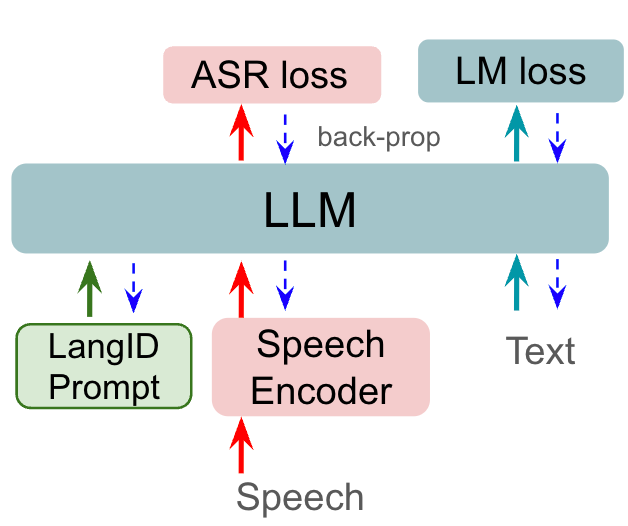}}
    \caption{Previous works [a] and [b] denotes the baseline prefixLM and soft prompting with prefixLM respectively. Subfigures [c] and [d] are our proposed approaches representing the prefix-tuning with RNNT loss and langID based soft prompting respectively}
    \label{fig:prev_works}
\end{figure}

\noindent Our primary contributions are as follows:
\begin{itemize}
    \item RNNT loss for speech prefix tuning: We demonstrate  improvements over both frozen and fine-tuned LLMs. We also examine the constraints encountered when tuning with speech prefixes using Connectionist Temporal Classification (CTC) loss, a non-autoregressive technique. We compare CTC with RNNT loss, highlighting distinctions relevant to prefixLM. 
    \item Language ID (langID) soft prompting: This technique enhances performance of frozen LLMs.
    \item Bridging the gap: Applying both speech prefix-tuning and langID-based soft prompting can be additive and further reduce the performance gap between frozen and fine-tuned LMs. 
\end{itemize}


%% file: inputfiles/methods.tex
\begin{figure}[!ht]
    \centering
    \includegraphics[width=0.9\linewidth]{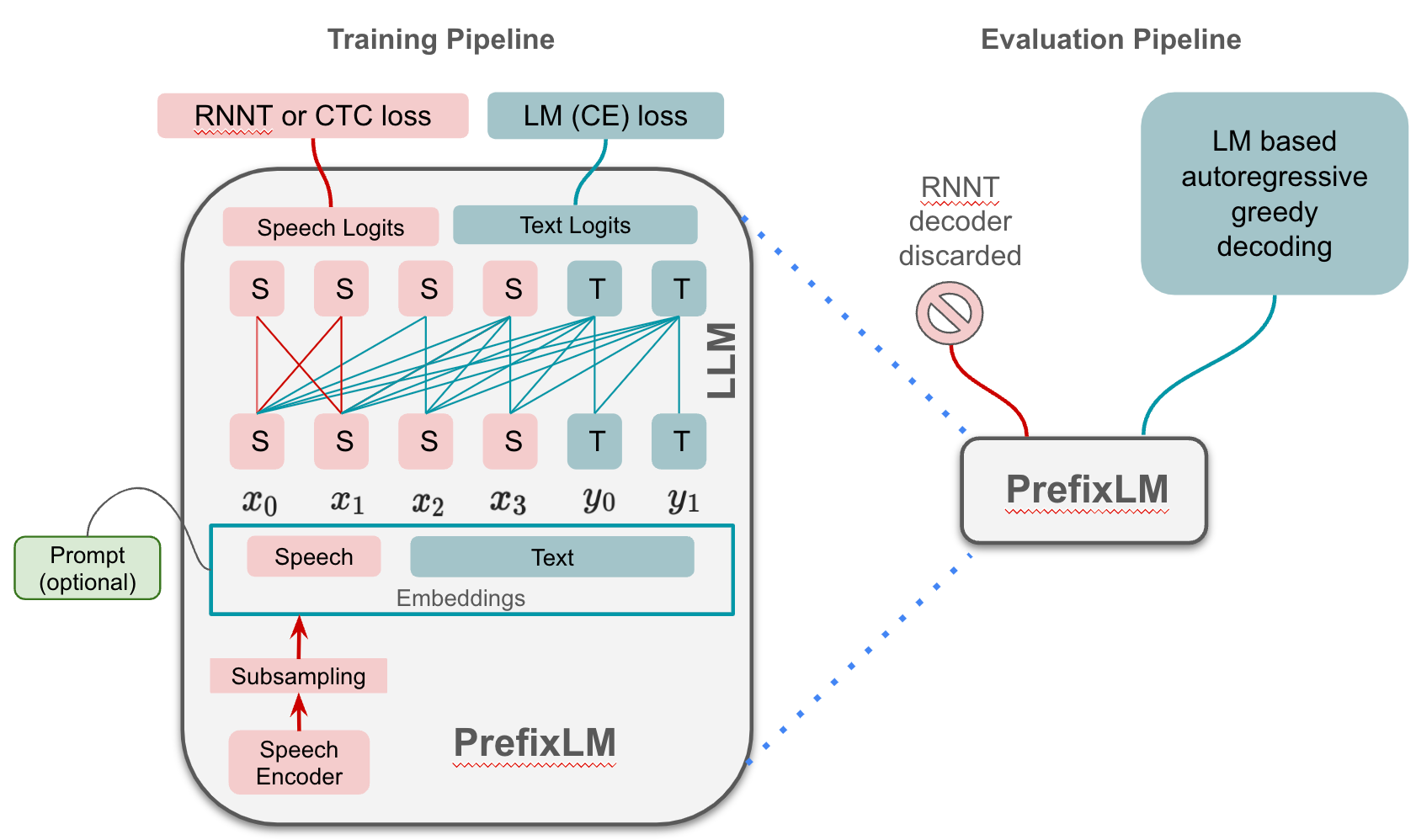}
    \caption{Training and Evaluation flow for PrefixLM with speech prefix-tuning}
    \label{fig:flowchart}
\end{figure}

Our proposed method focuses on finetuning the speech prefix tokens of prefixLM with ASR loss for improved recognition performance. Figures~\ref{fig:flowchart} presents the training and evaluation pipeline for our proposed speech prefix-tuning approach. Unlike previous works in figure~\ref{fig:prev_works} that solely focus on tuning only the prefix embeddings with the same loss used for text prediction, tuning with RNNT loss updates both speech encoder and prefix embeddings allows the model to learn more discriminant speech features as prefix tokens.

Given a input speech sequence $\mathbf{X}=x_{0},x_{1},x_{2},x_{3}$ with $T=4$ frames and text sequence $\mathbf{Y}=y_{0},y_{1}$ with $U=2$ length, the prefixLM $f(\cdot)$ takes the concatenated input $[X,\,Y]$.

\subsection{PrefixLM}
PrefixLM~\cite{raffel2020exploring} is a decoder only model operating in an input-to-target paradigm.  It can be viewed as almost encoder-decoder models with shared parameters. PrefixLM has shown competitive advantage~\cite{ding2023causallm} as an adaptation method for tasks with relatively small amount of data.
The PrefixLM architecture $f(.)$ intakes $[X,\,Y]$ and enables bi-directional attention on the prefix sequence $x_{0:T-1}$. This serves as the prefix for subsequent prediction on $x_{T:N}$.
The output logits $\hat{Z}_{0:T-1} = \hat{x}_{0:T-1}$ corresponding to the acoustic prefix are discarded and the output logits of predicted text $\hat{Z}_{T:T+\hat{U}-1} = \hat{y}_{0:\hat{U}-1}$ are used for decoding.
The final training objective to learn the model parameters $\phi$ is done minimizing the errors during next text token prediction using CE loss:
\begin{equation}
    \mathcal{L_{\mathrm{LM}}} = - \sum_{u=1}^{\hat{U}} \log p_{\phi} (\hat{y}_{u} \mid \hat{y}_{0:u-1}, \mathbf{X}).
    \label{eq:1}
\end{equation}
\subsection{Prefix-tuning}
Prefix-tuning is fine-tuning only the embedding layer which contains the stack of prefixes to guide the text embeddings towards the target task. 
During training with the equation~\eqref{eq:1}, the prefix embeddings learns the abstract representations of the underlying task. The prefix embeddings remain fixed during inference and projects the required information from  the evaluation dataset. Prefix-tuning is computationally efficient with much fewer trainable parameters and also avoids over-fitting.

\subsection{RNNT decoder}
RNNT~\cite{ghodsi2020rnn} is an autoregressive sequence-to-sequence model that processes the encoded speech sequence to generate a distribution of text tokens for each timestep of the encoder output. A joint network then combines encoder information and the previous prediction to generate the current token. The RNNT decoder relies on the output of a speech encoder. In our work, we propose using the output logits of a prefixLM's speech prefixes as input to the RNNT decoder.

\subsection{Speech prefix-tuning with RNNT loss}\label{sec:sptal}
Based on prior ASR works with LLM~\cite{embarassprefix,fathullah2023prompting,lakomkin2023end}, we believe that having well encoded speech prefixes act as better context to drive the LM towards the target task. The prefixLM has the ability to learn the speech-to-text alignment better as the speech sequence length gets closer to the text sequence length~\cite{fathullah2023prompting}. 
Extending this intuition beyond the sequence length, we want to find better speech representations that steers the LM to improve the ASR task. Intuitively, the speech prefixes can influence the text encodings $\mathbf{Y}$ by guiding what to extract from $\mathbf{Y}$; and can improve the text generation by driving the next token distribution. The proposed objective to use ASR loss amplifies the distinctiveness of the speech features by updating the speech related parameters.

To perform speech prefix-tuning, we propagate the speech output logits from LLM to the RNNT decoder. The RNNT loss learns the speech-to-text alignment and $\langle \rm{eos}\rangle$ prediction allowing the speech prefixes to accomodate more knowledge of underlying speech and the text to be predicted. The joint training objective is:
\begin{align}\label{eq:2}
    \mathcal{L}_{\mathrm{RNNT}} &= - \log p(\mathbf{Y} \mid \mathbf{\hat{X}}) \\ 
    \mathcal{L}_{\mathrm{joint}} &= \alpha \cdot \mathcal{L}_{\mathrm{LM}} + (1 - \alpha) \cdot \mathcal{L}_{\mathrm{RNNT}}  \label{eq:3}
\end{align}
Here, $\hat{\mathbf{X}}$ is the speech prefix output (speech logits) from the LLM as in figure~\ref{fig:flowchart}.
\subsection{Language ID based soft prompting}
\label{ssec:lid-prompt}
Prefix tuning learns task specific information and conditioning the prefix with LangID allows it to learn language specific embeddings. This helps to stabilize performance on multiple languages when the LLM is frozen. Figure~\ref{fig:prev_works}[d] shows the training and eval pipeline for langID based prompting. The LangID embeddings are of size $[L \times M \times D]$, where $L$ is the number of languages, $M$ is the prompt length and $D$ is the number of dimensions. The soft prompt $[M \times D]$ is chosen corresponding to the langID from the source input and is fed along with the speech prefixes from Section~\ref{sec:sptal}. During training, the soft prompt embeddings are only updated and are fixed during inference.
\subsection{LLM training}
LLMs can be trained in one of the following ways when using speech prefixes:
\begin{itemize}
    \item Frozen LLM: The RNNT loss updates only the speech encoder and the soft prompt embeddings while the entire LLM parameters $\phi$ are kept frozen. 
    The LM loss~\eqref{eq:1} updates the soft prompt embeddings only.
    \item Finetuned LLM: LLM parameters are updated simultaneously with both ASR and LM loss. The soft prompt embeddings are tuned only with the LM loss given in equation~\eqref{eq:1}.
\end{itemize}

%% file: inputfiles/experiments.tex
\subsection{Datasets and Models}
\noindent\textit{Data}: The training data used in these experiments is composed of YouTube longform data  as described in \cite{liao_2013_youtube, chen_2023_largescale} and drawn from 10 Indic languages. All data is drawn  from 10 Indic languages and segmented into ``utterances'' with a maximum length of 30s.
Language information is obtained from the uploaded language tag in the video and incorporated as an auxiliary embedding along with speech features. 
For evaluation we use a YouTube test set for 10 Indic languages which combines utterances spanning a broad set of topics ranging from sports and entertainment to education. Both training and test data is segmented into utterances with a maximum length of 30s.
See Table \ref{tab:datastats} for the distribution of training and test material across languages.
\begin{table}[!ht]
    \centering
    \caption{Training and testing data statistics}
    \resizebox{0.5\linewidth}{!}{
    \begin{tabular}{cccc}
    \toprule
         LID & Language & \multirow{1}{*}{\#Hours} & \multirow{1}{*}{\#Hours} \\
          &  & (Train) & (Test) \\ 
    \midrule
         bn & Bengali & 3.3k & 30.2 \\
         en & English & 3.5k & 22.2 \\
         gu & Gujarati & 3.5k & 30.4 \\
         hi & Hindi & 5.5k & 30.1 \\
         kn & Kannada & 3.6k & 29.8 \\
         ml & Malayalam & 3.2k & 29.3 \\
         mr & Marathi & 3.7k & 30.0 \\
         ta & Tamil & 4.5k & 28.7 \\
         te & Telugu & 4.2k & 29.6 \\
         ur & Urdu & 2.0k & 30.2 \\
    \bottomrule
    \end{tabular}
    }
    \label{tab:datastats}
\end{table}

\noindent\textit{Speech Encoder}: We employ universal speech models (USM)~\cite{zhang2023google} with model complexity of 300M (USM-S) and 600M (USM-L) parameters. USM-S leverages a 24-layer Conformer with a model dimension (768) resulting in a total of 333.5 million parameters while USM-L has the same number of layers as USM-S but with 1024  dimensions. Both USM architectures use chunk-wise bi-directional attention allowing them to accurately model long audio sequences (30-second segments during training).  Mel fiterbank based speech features are fed to the USM speech encoder and the encoded outputs are subsampled by factor of 4 (160ms frame rate) for efficiency. This subsampled encoder output $\mathbf{X}$ serves as the prefix embedding.
The USM is trained using a large amount of multilingual data: over 10 million hours of unlabeled audio, tens of billions of text sentences, over a hundred thousand hours of supervised and semi-supervised audio. The data is drawn from over a hundred languages covering various topics~\cite{zhang2023google}.

\noindent\textit{LLM}: The large language model used in this paper builds upon the JAX based M4 multipod model 
~\cite{chowdhery2023palm}). This a Transformer based decoder only model. 
In this paper, we present results with two LLM sizes, 128M and 500M parameters. 128M has 8 layers with 16 heads, 4096 hidden dimensions. 500M model has 30 layers with 16 heads and 4096 hidden dimensions. The feed-forward layer configuration is common to both 128M (LLM-S) and 500M (LLM-L) parameter models with 16384 dimensions and the attention head size is 64. Both these models are trained with 800B text tokens. 
We use relative positional embeddings and GELU activations. Adafactor optimizer with momentum is used for training with a batch size of  1024 and a sequence lengths of 1k tokens. Finally, the model is quantized to bfloat16 precision for efficient tuning and inference. 
256k vocab based sentencepiece tokens~\cite{kudo2018sentencepiece} are used for training. Training is performed using a variant of UL2 objective \cite{tay2022unifying}, 
as described in~\cite{garcia2023unreasonable}.





%% file: inputfiles/discussion.tex
\begin{table}[!ht]
    \centering
    \caption{WER on the average of 10 Indic languages using CTC, RNNT and PrefixLM using 300M (USM-S) and 600M  (USM-L) speech encoders. PrefixLM (finetuned) model is trained using $\mathcal{L}_{\mathrm{LM}}$ in ~\eqref{eq:1} and PrefixLM (prefix-tuned with RNNT) model is trained using $\mathcal{L}_{\mathrm{joint}}$ in ~\eqref{eq:3}.
    }
    \resizebox{0.99\linewidth}{!}{
    \begin{tabular}{cc|c|c}
    \toprule
         Train decoder & Eval decoder & \multicolumn{2}{c}{Avg WER (\%)}  \\
         &  &  USM-S & USM-L \\
         \midrule
         CTC &  CTC & 35.9 & 33.0 \\
         PrefixLM (finetuned) &  LM& 36.7 & 32.2 \\
         PrefixLM (prefix-tuned with CTC) &  CTC & 33.8 & 31.8 \\
         PrefixLM (prefix-tuned with CTC) & LM & 33.7 & 31.8 \\
         \midrule
         RNNT &  RNNT & 31.5 & 29.4 \\
         PrefixLM (prefix-tuned with RNNT) &  RNNT & 29.8 & 28.4 \\
         PrefixLM (prefix-tuned with RNNT) & LM & 29.8 & 28.3 \\         
         \bottomrule
    \end{tabular}
    }
    \label{tab:prefixlm_ctc_300m}
\end{table}
\begin{table*}[!ht]
    \centering
    \caption{WER on all 10 Indic languages using prompt tuning, prefix tuning and language ID prompt tuning with and without fine-tuning the LLM.
    }
    \resizebox{0.9\linewidth}{!}{
    \begin{tabular}{ccccccccccccc}
    \toprule
         LLM & Tune & bn & en & gu & hi & kn & ml &  mr & ta & te & ur & Avg\\
         \midrule
         \multirow{5}{*}{USM-L +  frozen LLM-L} & - & 33.5 & 16.6 & 51.1 & 49.4 & 57.6 & 54.6 & 30.3 & 52.2 & 45.7 & 45.7 & 43.6 \\
          & Prompt & 33.5 & 15.2 & 50.2 & 45.1 & 52.3 & 51.1 & 30.4 & 52.0 & 44.6 & 41.0  & 41.5 \\
            & Prefix & 22.0 & 14.1 & 37.8 & 15.5 & 37.6 & 40.1 & 27.3 & 42.0 & 33.0 & 21.4 & 29.1 \\
           & Prefix+Prompt & 20.9 & 14.5 & 37.6 & 15.3 & 37.4 & 39.7 & 26.9 & 42.2 & 32.7 & 21.3 & 28.9 \\
         & Prefix+LangIDPrompt & 20.5 & 14.3 & 37.0 & 15.2 & 37.2 & 39.4 & 26.4 & 41.1 & 32.4 & 21.0 & 28.5 \\
         \midrule
         \multirow{2}{*}{USM-L + finetuned LLM-L} & - & 27.5 & 17.4 & 40.7 & 18.3 & 40.6 & 42.8 & 29.9 & 44.0 & 35.6 & 25.6 & 32.2 \\
            & Prefix & 20.2 & 13.7 & 37.1 & 15.2 & 37.2 & 39.5 & 26.5 & 41.5 & 32.4 & 21.0 & 28.3 \\
         \midrule
    \end{tabular}
    }
    \label{tab:prefixlm_comparison}
\end{table*}

\subsection{PrefixLM with CTC and RNNT}\label{sec:ctcrnnt}

The introduction of RNNT or CTC decoders and corresponding losses to a PrefixLM ASR model demonstrates clear improvements.  Using a CTC auxiliary loss, WER on USM-L drops from 32.2 to 31.8 with a larger win on USM-S, where performance goes from 36.7 to 33.7 (8\% relative).  RNNT in isolation is a better ASR model than CTC (29.4 vs. 33.0) given by rows 1 and 5 in Table~\ref{tab:prefixlm_ctc_300m}. This is also reflected in its combination with PrefixLM.  The introduction of the RNNT decoder and loss to PrefixLM  yields 5.4 \% and 3.7\% relative wins on USM-S and USM-L respectively. 

\subsection{Language-based Prompting allows the LM to be frozen}
The results in Table \ref{tab:prefixlm_ctc_300m}  update the full PrefixLM model (Figure~\ref{fig:flowchart}), both the speech encoder and LM.  However, updating the LLM is computationally expensive. In this section we explore in-context, prompting techniques to achieve similar performance by updating the speech encoder while keeping the LLM frozen.

Table \ref{tab:prefixlm_comparison} also shows the results of using both learned soft-prompts and language-conditioned prompts as described in Section \ref{ssec:lid-prompt} using LLM-L with USM-L as the speech encoder. The average \%WER performance of using frozen LLM is significantly worse with 43.6 over finetuned LLM with 28.3. Prompt tuning shows that it improves on average by absolute 2.1\%. Our proposed speech prefix-tuning brings the average \%WER down to 29.1. Tuning both speech prefixes and prompt embeddings is complimentary and shows marginal gain.
We hypothesize this marginal improvement with speech prefix+prompt tuning is due to the limitation of using a single prompt embedding to model the multilingual input data. Extending the prompt to be language specific by using the langID prompt tuning is able to bring the performance of the frozen LLM to within 0.2\% WER of the best performing fully-updated model in Table \ref{tab:prefixlm_comparison} (28.5 vs. 28.3).  This results in a model where approximately half of the parameters do not need to be updated with a very modest impact on quality.

\subsection{Error analysis of speech prefix-tuning}
\begin{table}[!ht]
    \centering
    \resizebox{!}{0.5\linewidth}{
    \begin{tabular}{ccccc}
    \toprule
         \multirow{1}{*}{Lang} & \multirow{1}{*}{Error} & \multirow{1}{*}{RNNT} & \multirow{1}{*}{PrefixLM} & \multirow{1}{*}{PrefixLM} \\
          &  &  & (finetuned) & (prefix-tuned with RNNT) \\
    \midrule
          \multirow{3}{*}{bn} & D & 3.6  & \cellcolor{green!9}3.6 & \cellcolor{green!25}4.6 \\
                        &    I & 2.0 & \cellcolor{blue!25}{6.7} & \cellcolor{blue!9}{1.9} \\
                        &    S & 14.6 & \cellcolor{red!25}{17.2} & \cellcolor{red!9}{13.6} \\
    \midrule
        \multirow{3}{*}{en}   &    D & 3.1 & \cellcolor{green!9}3.0 & \cellcolor{green!25}3.6 \\
                        &    I & 2.7 & \cellcolor{blue!25}{5.4} & \cellcolor{blue!9}{2.3} \\
                        &    S & 11.6 & \cellcolor{red!25}{9.0} & \cellcolor{red!9}{7.8} \\
    \midrule                        
        \multirow{3}{*}{gu}   &    D & 5.4 & \cellcolor{green!9}5.5 & \cellcolor{green!9}5.5 \\
                        &    I & 8.8 & \cellcolor{blue!25}{11.7} & \cellcolor{blue!9}{8.6} \\
                        &    S & 23.7 & \cellcolor{red!25}{23.5} & \cellcolor{red!9}{23.0} \\
    \midrule                        
        \multirow{3}{*}{hi}   &    D & 3.8 & \cellcolor{green!25}3.5 & \cellcolor{green!9}3.1 \\
                        &    I & 2.2 & \cellcolor{blue!25}{3.8} & \cellcolor{blue!9}{2.1} \\
                        &    S & 23.7 & \cellcolor{red!25}{11.0} & \cellcolor{red!9}{10.0} \\
    \midrule                        
        \multirow{3}{*}{kn}   &    D & 5.8 & \cellcolor{green!25}5.7 & \cellcolor{green!9}5.6 \\
                        &    I & 4.1 & \cellcolor{blue!25}{6.7} & \cellcolor{blue!9}{4.3} \\
                        &    S & 27.8 & \cellcolor{red!25}{28.2} & \cellcolor{red!9}{27.3} \\
    \midrule                        
        \multirow{3}{*}{ml}   &    D & 5.3 & \cellcolor{green!25}6.1 & \cellcolor{green!9}5.3 \\
                        &    I & 7.4 & \cellcolor{blue!25}{9.3} & \cellcolor{blue!9}{7.2} \\
                        &    S & 27.2 & \cellcolor{red!25}{27.5} & \cellcolor{red!9}{26.1} \\
    \midrule                        
        \multirow{3}{*}{mr}   &    D & 5.4 & \cellcolor{green!25}6.0 & \cellcolor{green!9}6.5 \\
                        &    I & 2.6 & \cellcolor{blue!25}{5.5} & \cellcolor{blue!9}{2.4} \\
                        &    S & 18.7 & \cellcolor{red!25}{18.4} & \cellcolor{red!9}{17.6} \\
    \midrule                        
        \multirow{3}{*}{ta}   &    D & 6.1 & \cellcolor{green!9}5.4 & \cellcolor{green!25}5.5 \\
                        &    I & 5.5 & \cellcolor{blue!25}{7.6} & \cellcolor{blue!9}{5.7} \\
                        &    S & 31.0 & \cellcolor{red!25}{31.1} & \cellcolor{red!9}{30.3} \\
    \midrule                        
        \multirow{3}{*}{te}   &    D & 4.5 & \cellcolor{green!9}4.4 & \cellcolor{green!25}4.7 \\
                        &    I & 5.4 & \cellcolor{blue!25}{7.7} & \cellcolor{blue!9}{5.4} \\
                        &    S & 23.0 & \cellcolor{red!25}{23.5} & \cellcolor{red!9}{22.3} \\
    \midrule                        
        \multirow{3}{*}{ur}   &    D & 3.4 & \cellcolor{green!25}4.0 & \cellcolor{green!9}3.1 \\
                        &    I & 6.2 & \cellcolor{blue!25}{7.6} & \cellcolor{blue!9}{6.3} \\
                        &    S & 19.4 & \cellcolor{red!25}{13.9} & \cellcolor{red!9}{11.7} \\
    \bottomrule
          
    \end{tabular}
    }
    \caption{\colorbox{green!25}{{\bf D}eletion}/ \colorbox{blue!25}{{\bf I}nsertion}/ \colorbox{red!25}{{\bf S}ubstitution} rate across PrefixLM and RNNT for Indian languages. The errors are color coded as \colorbox{red!25}{higher errors} and \colorbox{red!9}{lower errors} between the finetuned prefixlm $\mathcal{L}_{\mathrm{LM}}$ as in~\eqref{eq:1}  and PrefixLM with speech prefix-tuning ($\mathcal{L}_{\mathrm{joint}}$) in~\eqref{eq:3}.}
    \label{tab:dis}
\end{table}
Table \ref{tab:dis} shows that the PrefixLM (finetuned) model demonstrates a substantially higher rate of insertions.This is due to the effect of hallucinations during decoding. Our proposed approach speech prefix-tuning with RNNT using $\mathbf{L}_{\mathrm{joint}}$ loss reduces this insertion rate while maintaining the overall quality. The average performance gains in table~\ref{tab:prefixlm_ctc_300m} are attributed primarily to improvement in insertions and substitutions without hurting the deletion rate. We observe this behavior for all 10 Indic languages.

\subsection{Code-switching analysis}
Without language ID information, multilingual ASR models have a tendency to produce hypotheses in multiple languages, sometimes multiple scripts.  Some of this is by design, as speech in Indian languages is frequently code mixed. However, this is also a source of error where a hypothesis may be acoustically ``correct'' but produced in an unexpected script.  Here we measure the code-mixing behavior of the different approaches using the Code Mixing Index (CMI) measure denoted in \cite{cmi}:

\begin{equation}
\mathrm{CMI} =
\begin{cases}
100 * [\frac{1 - \mathrm{max}(w_{i})}{n-u}]; & n>u \\
0\,\,\,\,\,\,\,\,\,\,\,\,\,\,\,\,\,\,\,\,\,\,\,\,\,\,\,\,\,\,\,\,\,\,\,\,\,\,\,\,; & n=u
\end{cases}
\end{equation}
Here, 
 $\mathrm{max}(w_{i})$ = highest number of words present from any language (more than 1 language can have the same highest word count), $n$ = no of tokens in utterance $x$, $u$ = number of tokens given other language tags.  
 
\begin{table}[!ht]
    \centering
        \caption{Code Mixing percentage on Tamil (ta) for finetuned prefixlm $\mathcal{L}_{\mathrm{LM}}$ and PrefixLM with speech prefix-tuning ($\mathcal{L}_{\mathrm{joint}}$) models}
    \label{tab:CMI}
    \resizebox{0.9\linewidth}{!}{
    \begin{tabular}{cccc}
    \toprule
         Model & \# words & \# maxwords in ta & CMI  \\
    \midrule
         RNNT & 137257 & 136788 & 34.2\% \\
         PrefixLM (finetuned) & 137394 & 136888 & 36.8\% \\
         Proposed  & 137299 & 136849 & 32.8\% \\
    \bottomrule
    \end{tabular}
    }
\end{table}

In Table \ref{tab:CMI} we see that, for Tamil (ta), the PrefixLM generates more code-mixed hypotheses than the proposed speech prefix-tuning $\mathcal{L}_{\mathrm{joint}}$ model. The \%CMI improves from 36.8 to 32.8 which shows that the number of non Tamil words are predicted less compared to the baseline.  
 This behavior was observed across other Indic languages as well.

%% file: inputfiles/related.tex
In just the last few years there has been a lot of work on using LLMs within ASR.  These include Flamingo \cite{alayrac2022flamingo}, PrefixLM \cite{raffel2020exploring}, and SLM \cite{wang2023slm}. Some other notable works~\cite{chen2023lauragpt,zhang2023speechgpt,yu2023connecting}, merge pretrained speech encoder with pretrained text based LLM to perform ASR and other speech related tasks. Given the rate of progress this is a necessarily incomplete list. Most of these works rely on a well pretrained speech encoder~\cite{zhang2023google} with matching domain to achieve better recognition performance. On the other hand, the LLMs are adapted to the target domain by performing either complete finetuning~\cite{wu2023decoder,tang2023salmonn} or other lightweight approaches such as prompt-tuning~\cite{yu2024prompt, embarassprefix,prompttuningv2} . 

In this work, we show that these two techniques can be unified by performing speech prefix-tuning using a joint RNNT and LM loss provides both better prefix embeddings and also performs lightweight finetuning. To further reduce the tunable parameters, we present the langID based soft prompting in Section \ref{ssec:lid-prompt} by using conditioning information to learn a soft-prompt, rather than hand crafting a hard-prompt for this conditioning. 





%% file: inputfiles/conclusion.tex
The inclusion of traditional RNNT loss successfully complements the success of PrefixLM-based ASR.  The overall quality remains high, while balancing the insertion rate.  Moreover the rate of code-mixed output is reduced.  We have also demonstrated the value of learned soft-prompts conditioned on language ID as a route to eliminate the need for a fine-tuned LM, substantially reducing the training required for this technique to obtain high quality results.